# Automated Detection of Acute Promyelocytic Leukemia in Blood Films and Bone Marrow Aspirates with Annotation-free Deep Learning

Petru Manescu[1*], Priya Narayanan[1], Christopher Bendkowski[1], Muna Elmi[1], Remy Claveau[1], Vijay Pawar[1], Biobele J. Brown[2], Mike Shaw[1], Anupama Rao[3], and Delmiro Fernandez-Reyes[1,2*]

1. Dept. of Computer Science, Faculty of Engineering, University College London, Gower Street, London, UK.

2. Dept. of Paediatrics, College of Medicine University of Ibadan, University College Hospital, Ibadan, Nigeria.

3. Dept. of Haematology, Great Ormond Street Hospital for Children, London, UK.

**\*Corresponding Authors:**

Dr. P. Manescu (p.manescu@ucl.ac.uk); Prof. D. Fernandez-Reyes (delmiro.fernandez-reyes@ucl.ac.uk).

**ABSTRACT**

While optical microscopy inspection of blood films and bone marrow aspirates by a hematologist is a crucial step in establishing diagnosis of acute leukemia, especially in low-resource settings where other diagnostic modalities might not be available, the task remains time-consuming and prone to human inconsistencies. This has an impact especially in cases of Acute Promyelocytic Leukemia (APL) that require urgent treatment. Integration of automated computational hematopathology into clinical workflows can improve the throughput of these services and reduce cognitive human error. However, a major bottleneck in deploying such systems is a lack of sufficient cell morphological object-labels annotations to train deep learning models. We overcome this by leveraging patient diagnostic labels to train weakly-supervised models that detect different types of acute leukemia. We introduce a deep learning approach, Multiple Instance Learning for Leukocyte Identification (MILLIE), able to perform automated reliable analysis of blood films with minimal supervision. Without being trained to classify individual cells, MILLIE differentiates between acute lymphoblastic and myeloblastic leukemia in blood films. More importantly, MILLIE detects APL in blood films (AUC 0.94±0.04) and in bone marrow aspirates (AUC 0.99±0.01). MILLIE is a viable solution to augment the throughput of clinical pathways that require assessment of blood film microscopy.

**INTRODUCTION**

Morphological evaluation of leukocytes from peripheral blood smears and bone marrow aspirates under a high numerical aperture objective is an important step in diagnosing hematopoietic malignancies such as acute leukemia[1]. More specifically, blood smears must always be inspected in the case of unexplained leukocytosis or when a complementary automated instrument suggests the presence of blasts[2]. Equally, this allows differentiating between myeloid and lymphoid lineages which is crucial for treatment selection[3–5].

Unfortunately, examination of peripheral blood and bone marrow aspirate films strongly relies on the availability of trained personnel, is time-consuming and prone to human error due to fatigue and cognitive overload. The emergence of digital pathology has presented the potential for scalable artificial intelligence assisted examination of peripheral blood films and bone marrow aspirates for diagnostic decision support[6]. Although computational hematopathology has shown potential in reproducing hematologists' work by training state-of-the-art supervised deep learning models to recognize well established morphological indicators of leukemia[7–14], a critical limitation of previous studies is that they do not focus on differentiating the type of leukemia such as Acute Lymphoblastic Leukemia (ALL) vs. Acute Myeloid Leukemia (AML). Furthermore, these studies have not attempted to detect cases of Acute Promyelocytic Leukemia (APL) which warrants emergency treatment impacting early mortality and prognosis[15] while other time-consuming parts of the clinical pathway are ongoing if available (e.g. genetics, cytochemistry, flow cytometry). An equally important drawback of previous fully supervised models[7,9,16,17] is that they require hundreds of thousands of object-level cell annotations provided by human experts[18] which are not only difficult to obtain at scale but are also susceptible to inconsistencies due to the subjectivity and cognitive fatigue of annotators. To overcome these limitations and, to provide a clinically relevant system that could support, together with clinical evaluation and ancillary laboratory parameters, the prompt treatment in APL cases, we designed a Multiple Instance Learning for Leukocyte Identification (MILLIE) approach. Our customizable annotation-free deep-learning framework leverages patient diagnostic labels to train weakly supervised models that detect different types of acute leukemia. While weakly supervised multiple instance learning models [19,20] trained with diagnostic labels have been previously used to analyze microscopy images in cell biology[21,22] and in computational cancer histopathology[23–26], there have been little attempts to apply its capabilities to the acute leukemia clinical pathway. Our results show that, despite not having been trained to classify individual cells, MILLIE can accurately distinguish between normal, acute lymphoblastic leukemia and acute myeloblastic leukemia by recognizing normal WBC, lymphoblasts and immature myeloid cells in peripheral blood films. MILLIE was equally able to distinguish AML bone marrow aspirates from healthy ones. MILLIE was also able to detect promyelocytes in both blood films and bone marrow aspirates as an indicator of acute promyelocytic leukemia (APL).

**METHODS**

For each of the experiments described above, we (1) extracted patches of individual white blood cells from each sample, (2) trained a multiple instance learning convolutional neural network model with sample-level labels and (3) applied the trained model and reported the results.

**Image Datasets**

All the datasets used in this study are publicly available. Table 1 summarizes the provenance and characteristics of the image datasets with diagnostic level labels (also termed sample level labels) used to train and test MILLIE. Table 2 summarizes the provenance and characteristics of the image datasets with single cell morphological annotations (also termed object level level) used to validate the multi-instance weakly supervised MILLIE approach.

**Image segmentation**

Images in RGB color space were converted to HSV space and Otsu's thresholding was applied to the Saturation channel as it offers high contrast stained WBC nuclei[27]. Morphological binary opening followed by watershed and removal of small blobs were further applied to segment WBC nuclei. Tiles of 200x200 pixels around the centroid of each remaining binary blob were cropped from the initial RGB images. We evaluated the segmentation accuracy on 15 randomly selected fields of view comprising a total of 71 manually annotated WBC. The segmentation algorithm "missed" 4 cells (recall:0.94) and detected 6 false positives (precision: 0.92).

**Network architecture and training details**

The convolutional layers of MILLIE models were initialized with weights from a VGG-19 model[28] pre-trained on the ImageNet dataset[29]. Objects of interest ($O_k^i$) with i=1,..,N corresponding to blood cells or parasite-like objects were extracted from each sample $k$ as p described in the previous section. MILLIE was trained to classify "bags" of these objects of interest with sample-level labels ($L_k$) provided by the routine clinical tests. The convolutional feature vectors corresponding to each cropped image ($F_k^i = conv(O_k^i)$) of the model were pooled into a single feature vector followed by two fully connected (FC) layers and a classification layer:

$$L_k = \mathbf{softmax}(\ldots(\mathbf{ReLu}(f_{fusion}(F_k^1, F_k^2,\ldots, F_k^N) \cdot W_1+b_1) \cdot \ldots \cdot)W_3+b_3) \quad (1)$$

where $N$ is the number of input image patches, $W_j, b_j$ are the corresponding weights and biases of each FC layer and $f_{fusion}$ is the feature aggregation rule. More specifically:

$$f_{fusion}^{i=1\ldots N}(F_k^i) = \max_{1\leq i \leq N}(\zeta_{l=1\ldots n_f}^i) \quad (2)$$

with $n_f$, the number of individual features ($\zeta_l^i$) in each $F_k^i$.

Up to fifty image patches subject to on-the-fly geometrical augmentation (random rotations and random flips) as well as spectral augmentation (random hue modification, random gamma corrections, random noise) were randomly selected per sample for each iteration during training. The same augmentations were applied during testing for both sample-level and cell-level prediction. In this way, the spectral augmentations compensate for any covariate (acquisition) shifts due to the differences in the cameras and microscope settings used to image the different datasets. We employed stochastic gradient descent with a learning rate of 0.0003 and a cross entropy loss function to optimize the model weights during maximum 100 epochs (or early stopping). At testing time, all image patches from each sample were passed through the network.

**RESULTS**

We trained MILLIE models to distinguish between normal, ALL and AML samples and we evaluated their classification performance both at sample and cell-level on separate hold out test sets. Our approach extracts patches containing individual cells from high-resolution fields of view of peripheral blood films and bone marrow aspirates (Fig. 1a) and employs these patches to train a weakly supervised convolutional neural network model with diagnostic labels[30] (Fig. 1b). More specifically, MILLIE was trained to differentiate between "bags" of cell instances extracted from positive samples (containing both regular and abnormal white blood cells) and "bags" of cell instances extracted from negative samples (only regular cells). Trained to predict patient diagnosis, MILLIE implicitly identifies disease specific indicators with minimum supervision (Fig. 1c) which renders the approach both highly efficient (e.g., not hindered by the lack and biases of object-level annotations) and interpretable.

**MILLIE differentiates between ALL and AML by recognizing myeloblasts and lymphoblasts in peripheral blood films**

We trained and validated our MILLIE model to predict diagnostic labels with image fields from 69 normal samples and 57 ALL samples publicly available[31] and 63 AML samples from a different publicly available databank (https://imagebank.hematology.org/). This combined dataset was randomly split into train (2/3) and test (1/3). MILLIE was then trained with "bags" of patches encompassing 200x200 pixels (12.8 µm x 12.8 µm) around the centroid of each previously segmented White Blood Cell (WBC)[27]. A 3-fold cross validation yielded an average accuracy over all three classes: 0.99±0.01 (Fig. 2d, confusion matrix for a single fold).

We further tested the model's ability to classify single cells on a separate publicly available dataset consisting of 130 images of individual normal WBC and 130 images of individual lymphoblasts[31] completed with 130 images of individual myeloblasts and immature myeloid cells randomly selected from another different public dataset[32]. We show that, despite being only trained on diagnosis level labels, MILLIE was able to recognize objects such as lymphoblasts (AUC=0.97) and myeloblast (AUC=0.97) cells with high accuracy (Fig 2 d-e). As a manner of interpretability, we investigated the cell-level feature space learned by MILLIE. The convolutional feature vectors of the individual test cell images were reduced to two-dimensional space for visualization through transformation via Principal Component Analysis (PCA) and each point was shaded by its ground-truth cell label (Fig. 2 g). PCA shows three distinct clusters of points with little overlap between normal and blast cells (Fig. 2 g).

**MILLIE recognizes acute promyelocytic leukemia and identifies promyelocytes in peripheral blood films**

We next tested MILLIE's ability to recognize APL samples and detect promyelocytes in blood films. For this purpose, we trained and validated a binary weakly supervised model to distinguish between APL (30 samples) and other (40 normal and other AML samples). Similarly, to the previous experiment, a random split (train:2/3 and test:1/3) and a 3-fold cross validation were performed. In terms of sample classification, MILLIE achieved an AUC of 0.935±0.036 (Fig 3c). Once trained and validated on weak sample-level labels, we further tested MILLIE's ability to distinguish promyelocytes from other types of WBC. On a separate test set comprising of single cell images of 611 promyelocytes and 3000 other myeloid and normal WBC randomly selected from a separate publicly available dataset[33], MILLIE achieved an AUC of 0.88 (Fig 3d). PCA of the learned convolutional features shows a distinct cluster corresponding to the promyelocytes slightly overlapping with the other types of cells (Fig. 3f). The fact that MILLIE is able to differentiate between images of promyelocytes and other types of WBC from a different public dataset that it was trained on suggests that the spectral augmentations during training corrected for any covariate shift caused by color variations between datasets.

To demonstrate the robustness of the method, we further trained and tested MILLIE to distinguish APL samples (n=33) from AML samples (n=72) from a different publicly available dataset [17]. MILLIE achieved an AUC of 0.96±0.02 (Fig 4a) on a random 3-fold cross validation and a higher AUC (0.94) than the one previously reported (0.86)[17]

**MILLIE detects AML and APL while identifying promyelocytes in bone marrow aspirates**

To further validate our approach, we trained and validated MILLIE to classify blood films of bone marrow aspirates from 236 healthy subjects and 1095 AML patients (out of which 43 were diagnosed with APL) using a publicly available dataset[16]. Random splits (train:3/4 and test:1/4) and a 4-fold cross validation were performed in this three-class classification problem. In terms of both AML and APL sample classification, MILLIE achieved on average an AUC of 0.99 (Fig. 5a). The confusion matrix (Fig. 5b) confirms the high accuracy in sample classification. Similar to our previous experiments, once trained and validated on diagnosis level labels, we further tested MILLIE's ability to distinguish myeloid immature cells (myeloblasts, monoblasts and promyelocytes) from mature healthy WBC (lymphocytes, monocytes and granulocytes) in bone marrow aspirates. On a publicly available test set comprising of manually annotated single cell images[16] of 309 promyelocytes, 718 myeloblasts and 262 normal WBC extracted from unseen samples, MILLIE achieved an AUC of 0.895 for promyelocytes and 0.862 on myeloblast classification (Fig 5c). While most promyelocytes (78%) and myeloblasts (84%) are classified correctly according to the weak labels, a fraction of the normal mature cells (mostly lymphocytes) is wrongly highlighted by MILLIE as AML cells (Fig. 5d). PCA of the learned features in the first fully connected layer shows three slightly overlapping but distinct clusters (Fig. 5e).

**DISCUSSION**

Our approach addresses a challenge in clinical automated computational hematopathology, namely detecting and differentiating between various types of immature white cells in peripheral blood films and bone marrow aspirates. Precise APL diagnosis relies on clinical suspicion, morphology, flow cytometry, and cytogenetic or molecular detection of translocation t(15;17)(q24;q21) PML-RAR which are time consuming if available at all in low resource settings. Furthermore, limited access to care, diagnostic delays leading to delays in the administration of all-trans retinoic acid (ATRA) and arsenic trioxide (ATO) therapy are all factors influencing the outcomes of patients with APL[34–36]. In this context, integrated within blood film and bone marrow assessment clinical workflows, deep learning computational pathology systems such as MILLIE could facilitate prioritization of acute leukemia diagnosis in currently overloaded healthcare systems as well as playing an important role in low-resource settings.

MILLIE produces high-performance interpretable deep-learning models for blood film analysis which are trained with diagnostic labels only, without any additional human-expert cell-level morphological annotations. Specifically, to train MILLIE we leveraged weak patient-level diagnosis labels to overcome the lack of annotations needed to train fully supervised machine learning models for white blood cell identification and classification. Despite being trained on such diagnosis labels (weak labels), MILLIE was successful in identifying individual well-established indicators associated with different types of acute leukemia, namely lymphoblasts, myeloblasts and promyelocytes. Detecting promyelocytes as an indicator of APL is extremely useful, as this supports emergency treatment which impacts patient prognosis[5].

We have also shown that MILLIE is adaptable and generally applicable to multi-class classification problems along with the binary diseased versus healthy classification tasks commonly examined in weakly supervised clinical contexts. Our computational hematology approach based on weak supervision is more appropriate for integration within clinical workflow than previous fully supervised approaches[16,18] since it only requires patient diagnosis level labels and does not rely on hard to obtain object-level labels. Equally important, our approach is more amenable to be deployed across different care centers as it allows each center to further train the system with specimens processed in their own facilities without the need to obtain large numbers of costly object level labels. Further studies could provide deep representations of patient level clinical history and diagnostic modalities (cytogenetic, molecular, flow-cytometry) that can be harnessed by our MILLIE approach to widen its image detection capabilities of reactive processes.

APL is a curable malignancy when appropriate prompt treatment is commenced. If deployed within overloaded hematology care pathways[15] our MILLIE computational hematology approach could transform the throughput by which blood films and bone marrow aspirates are assessed which could lead to prompt referral for treatment to reduce early mortality and improve prognosis of APL cases[15]. Regardless of the resource settings, MILLIE provides a realizable solution for clinical decision support and clinical pathway prioritization. In the context of remote resource poor healthcare settings, where added to the absence of cytogenetic or molecular testing capabilities, there is also lack of expertise for making the distinction between normal and abnormal blood films and marrow aspirates, MILLIE can provide decision support to initiate treatment. At the other extreme, in large urban resource rich countries, where there is large volume of patients across many complex clinical pathways, there is the advantage of rapid throughput to referral to effective treatment while reducing errors due to cognitive load of over stretched staff. Further studies implementing the MILLIE platform should allow the system to include a wider range morphological representations to improve throughput and accuracy hematological clinical pathways that require microscopic assessment of blood films or bone marrow specimens.

**Competing Interests**

The authors have declared that no competing interests exist.

**Data Availability Statement**

All the datasets used in this study are publicly available.

- Images from the ALL samples and corresponding labels are available from the ALL Image DataBase (ALL-IDB)[31] (http://homes.di.unimi.it/scotti/all/).
- Images from the AML samples (including APL) are publicly available at the American Society of Hematology Image Bank (https://imagebank.hematology.org/).
- Images of individual cells (promyelocytes and myelocytes) on which MILLIE was additionally tested are publicly available[33] at: https://data.mendeley.com/datasets/snkd93bnjr/1.
- Images and annotations of the bone marrow aspirate samples[16] are publicly available at:

https://www.kaggle.com/sebastianriechert/bone-marrow-slides-for-leukemia-prediction.

**Author Contributions**

DFR and PM designed the study. PM, MS, ME, RC, VP, CB and DFR carried out the experimental analysis. MS, PM, PN, RC, and CB developed the image pre-processing and segmentation software. DFR, CB, PM, ME, BJB, AR carried out laboratory experimental work. PM designed and developed the machine learning platform and its code. PM, DFR, BJB analyzed the data. PM and DFR prepared the manuscript with contributions from all authors. DFR is project lead. PM and DFR are corresponding authors.

**Table 1.** Image datasets with diagnostic level labels used to train and test MILLIE.

| Dataset Name | Ref. | Type | Website | Number of Samples/Image Fields | | | | |
|---|---|---|---|---|---|---|---|---|
| | | | | NORMAL | ALL | AML | APL | Total |
| ALL-IDB1 | 29 | PBS | http://homes.di.unimi.it/scotti/all/ | 69 | 57 | - | - | 126 |
| N/A | N/A | PBS | https://imagebank.hematology.org | - | - | 33 | 30 | 63 |
| N/A | REF | PBS | https://www.kaggle.com/eugeneshenderov/acute-promyelocytic-leukemia-apl. | - | - | 72 | 33 | 105 |
| N/A | 13 | BMA | https://www.kaggle.com/sebastianriechert/bone-marrow-slides-for-leukemia-prediction. | 236 | - | 1052 | 43 | 1331 |

PBS = peripheral blood film; BMA = bone marrow aspirate

**Table 2.** Image datasets with single cell (object-level) annotations used to validate MILLIE.

| Dataset Name | Ref | Type | Website | Annotated Cells (objects) | | | | |
|---|---|---|---|---|---|---|---|---|
| | | | | NORMAL | LYB | MYE | PMY | Total |
| ALL-IDB2 | 29 | PBS | http://homes.di.unimi.it/scotti/all/ | 130 | 130 | - | - | 260 |
| N/A | 30 | PBS | https://data.mendeley.com/datasets/snkd93bnjr/draft?a=d9582c71-9af0-4e59-9062-df30df05a121 | 10298 | - | 2284 | 611 | 13193 |
| N/A | 13 | BMA | https://www.kaggle.com/sebastianriechert/bone-marrow-slides-for-leukemia-prediction. | 844 | - | 4669 | 1038 | 6551 |

PBS = peripheral blood film; BMA = bone marrow aspirate; LYB = lymphoblasts; MYE = myeloid lineage cells; PMY = promyelocytes.

**Figure Legends:**

**Figure 1. MILLIE approach.**

**a.** Preprocessing and training data generation from a peripheral blood smear
Fields of view acquired with a high-NA objective lens (100x/1.4NA for blood films and 50x/0.55 for BMA) digital microscope are processed. A histogram-based segmentation[27] (see Methods section) was employed to generate binary masks corresponding to individual cells from the RGB images. These masks are further employed to crop individual patch images around each individual cell from the RGB images.

**b.** Training with weak labels. The extracted patches are passed through the convolutional neural network. Corresponding convolutional feature vectors are pooled together in one single feature vector (max pooling) followed by fully connected and classification layers. Weights of the model are optimized to predict the sample-level label available from routine clinical examinations.

**c.** Detecting morphological indicators. Once trained, individual cells can be passed one-by-one through the MILLIE models which classifies them as indicators for the specific disorders MILLIE learned to predict at a sample level.

**Figure 2. Acute leukemia detection and typing in blood films.**

**a-b**. Normal WBC and lymphoblasts detected in two ALL positive sample from the hold out validation set. **c.** Normal leucocytes and myeloblasts detected by MILLIE in an AML sample.

**d**. Confusion matrix for sample classification on the validation set.

**e.** Receiver operating characteristic (ROC) curve for single cell classification on the cell image test set.

**f.** Confusion matrix for cell classification (lymphoblasts vs myeloblasts vs normal) on the cell image test set.

**g.** PCA visualization of the convolutional representations learned by MILLIE of the individual cells in the test set.

**Figure 3. Acute promyelocytic leukemia detection in blood films.**

**a**. Promyelocytes detected in an APL positive sample from the hold out validation set.

**b**. Confusion matrix for sample classification on the validation set**.**

**c.** Receiver operating characteristic (ROC) curve for sample-level classification on the hold out validation set

**d.** ROC curve for cell-level classification on the cell image test set.

**e.** Confusion matrix for cell classification (promyelocytes vs other) on the cell image test set.

**f.** PCA visualization of the convolutional representations learned by MILLIE of the individual cells in the test set.

**Figure 4. APL vs AML classification of blood samples (additional dataset[17]).**

**a.** Receiver operating characteristic (ROC) curve for sample-level classification on the hold out validation set

**b**. Confusion matrix for sample classification on the validation set**.**

**c.** ROC curve for cell-level classification on the cell image test set.
**d.** Confusion matrix for cell classification (promyelocytes vs other) on the cell image test set.
**e.** PCA visualization of the convolutional representations learned by MILLIE of the individual cells in the test set.

**Figure 5. Acute myeloid leukemia detection in bone marrow aspirates.**
**a**. Receiver operating characteristic (ROC) curve for sample-level classification on the hold out validation set.
**b**. Confusion matrix for sample classification on the validation set**.**
**c.** ROC curve for cell-level classification on the cell image test set.
**d.** Confusion matrix for cell classification (promyelocytes vs other) on the cell image test set.
**e.** PCA visualization of the convolutional representations learned by MILLIE of the individual cells in the test set.

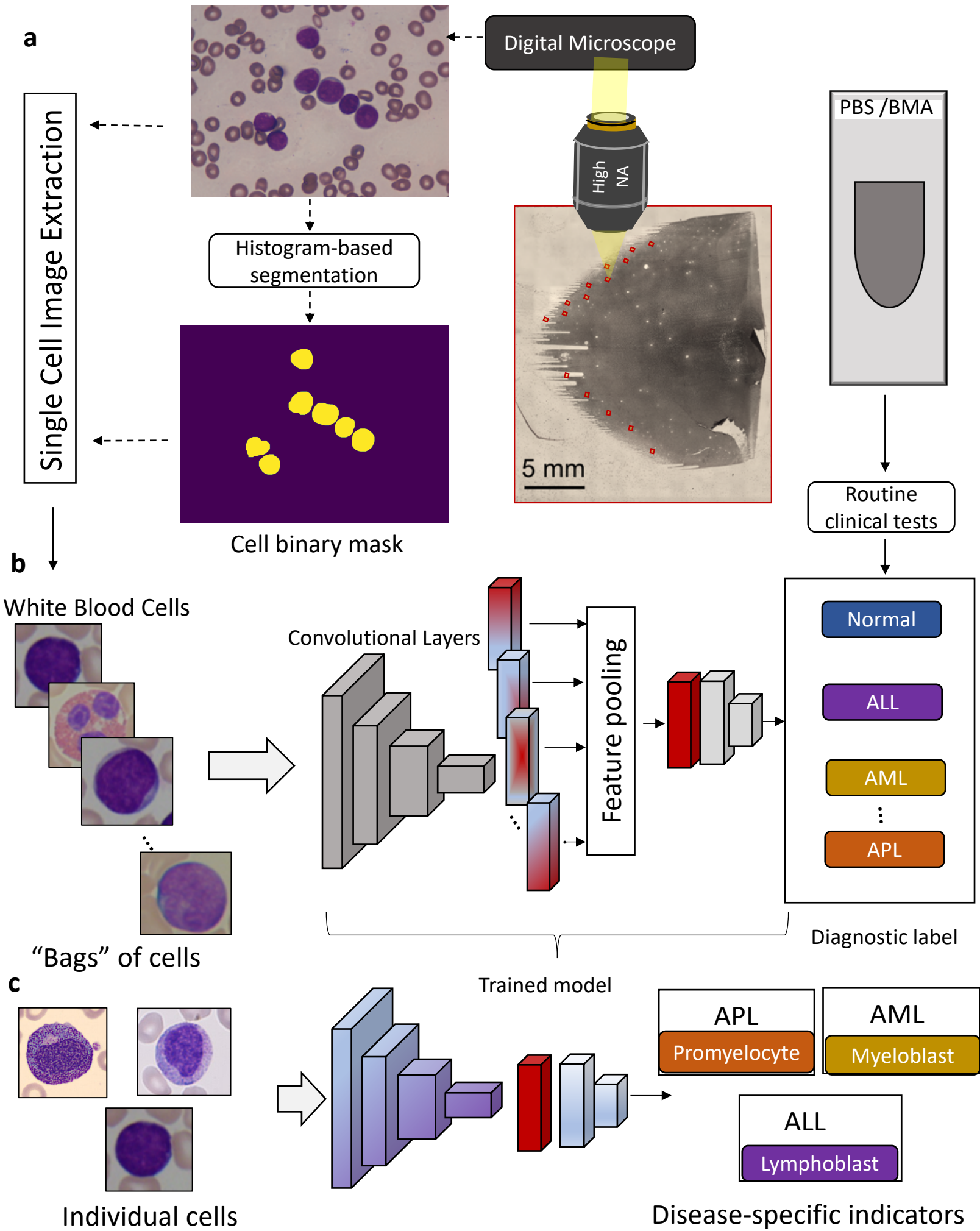

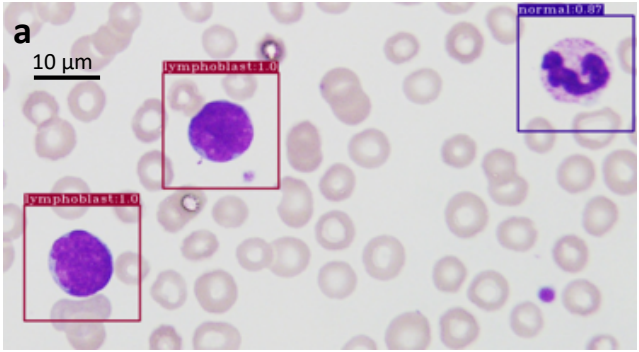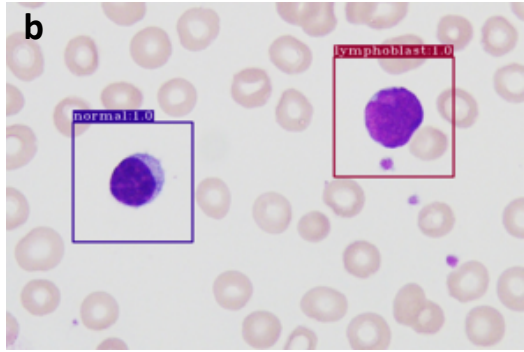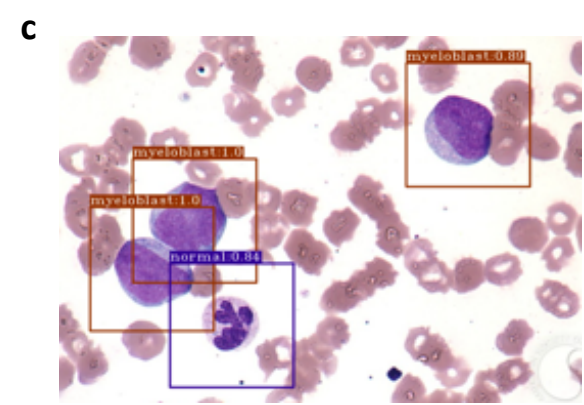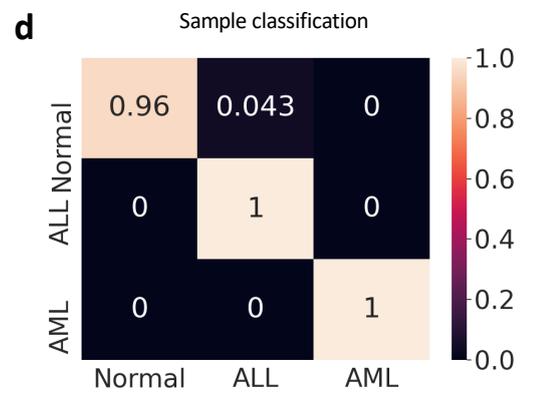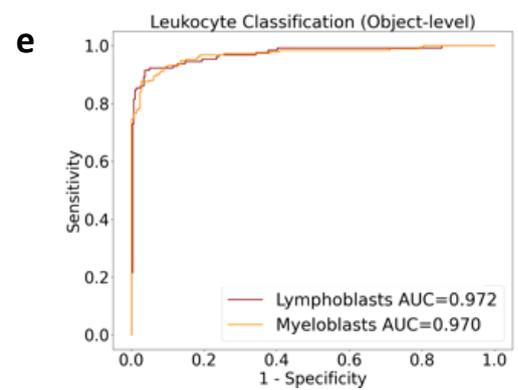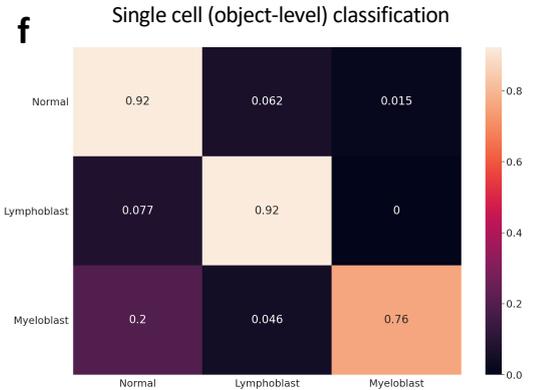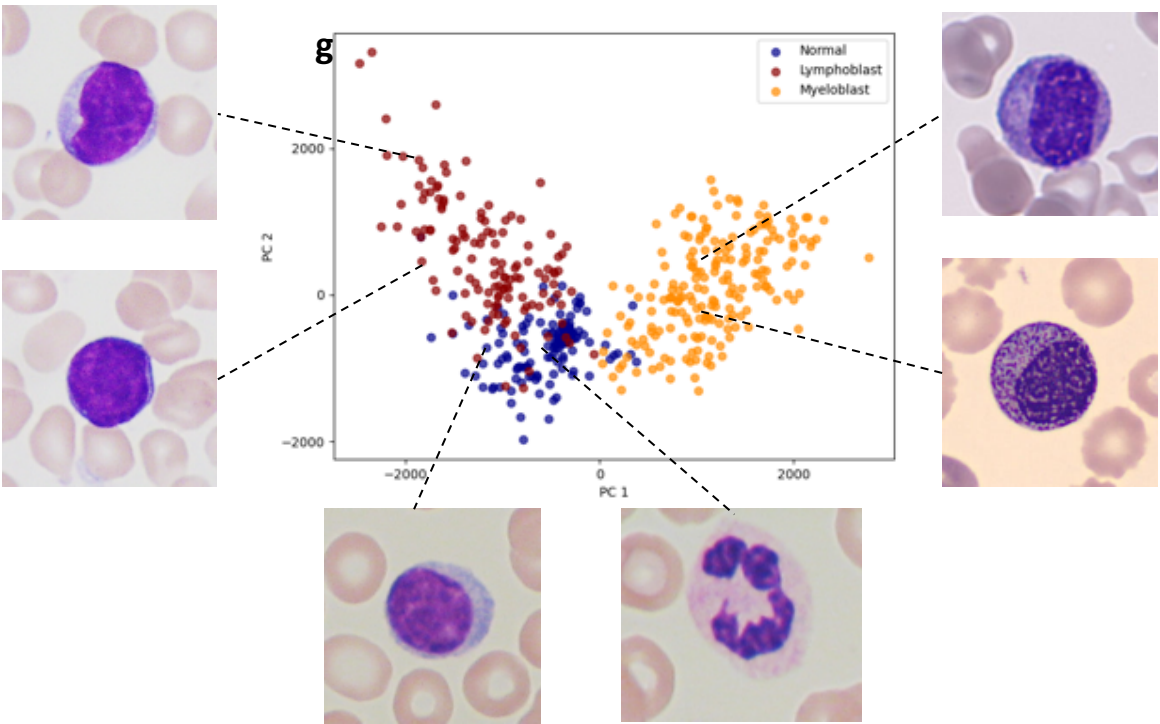

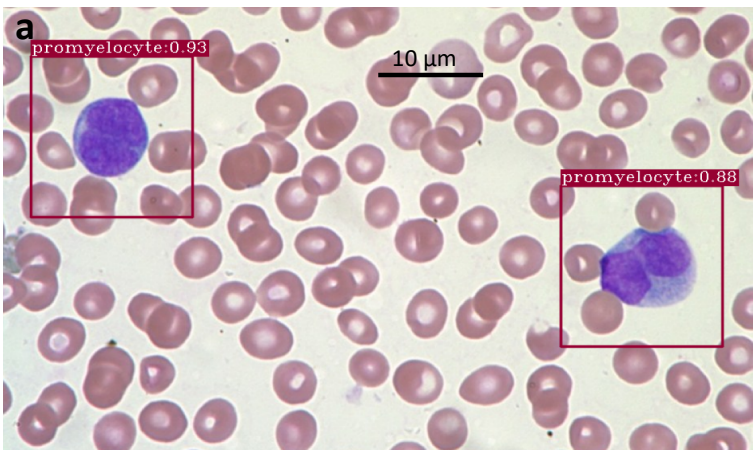
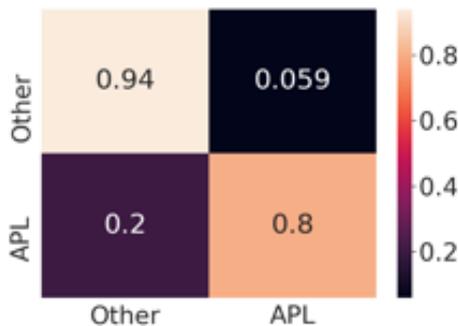
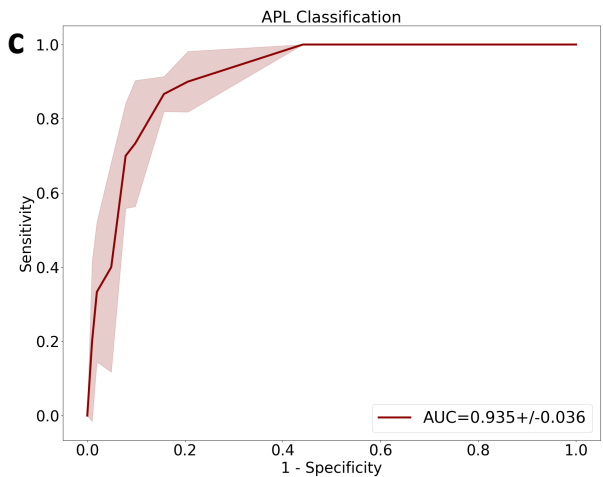
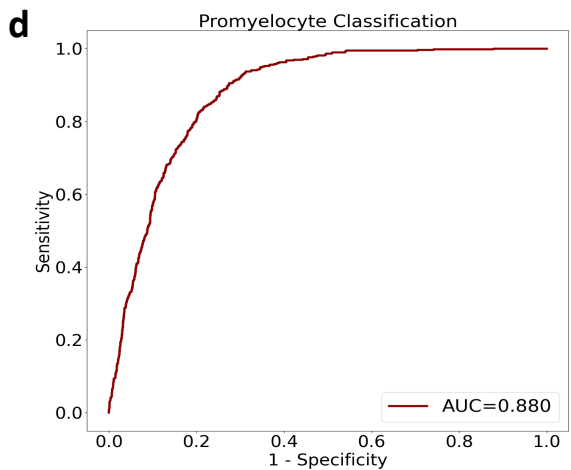
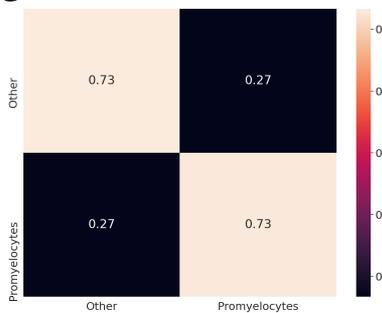
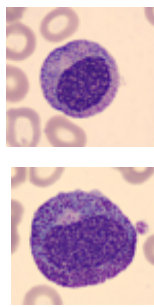
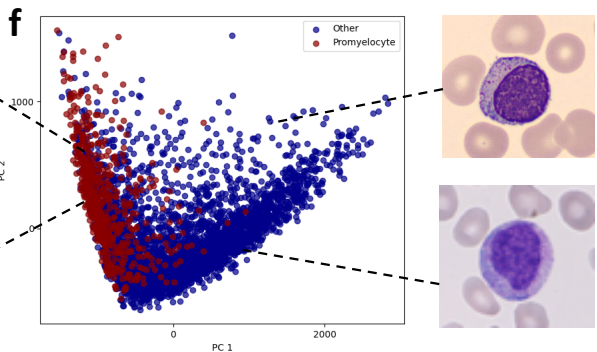

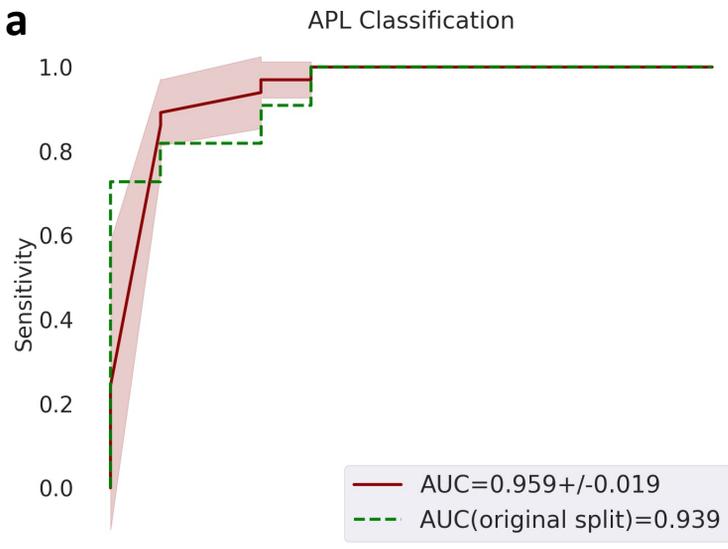
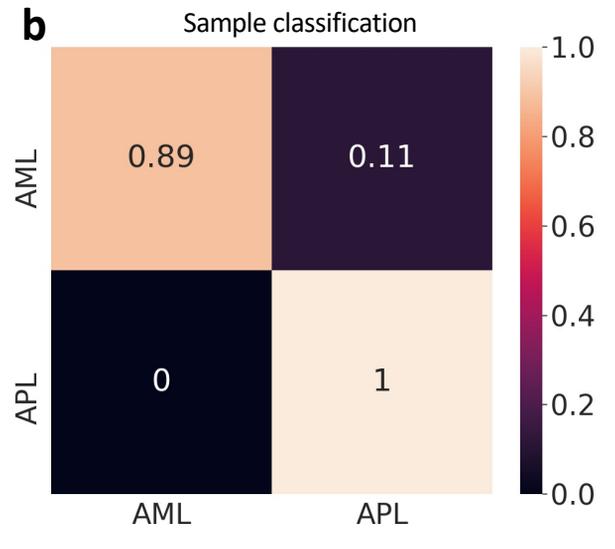
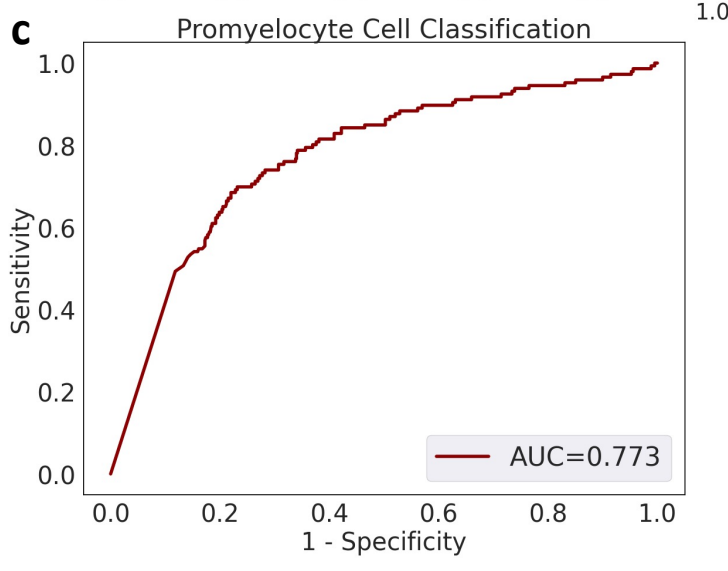
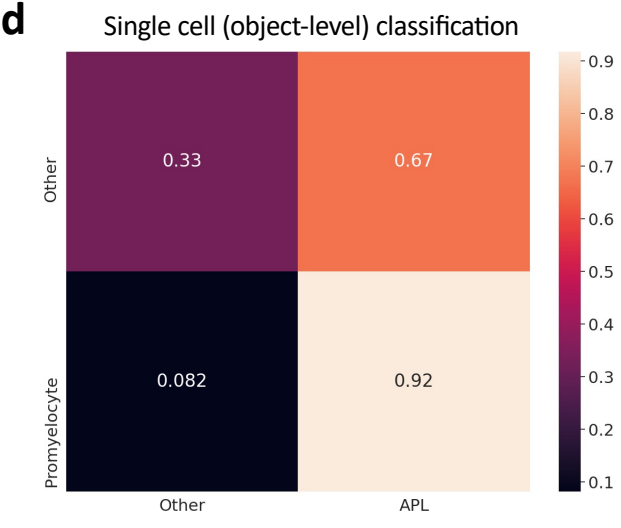
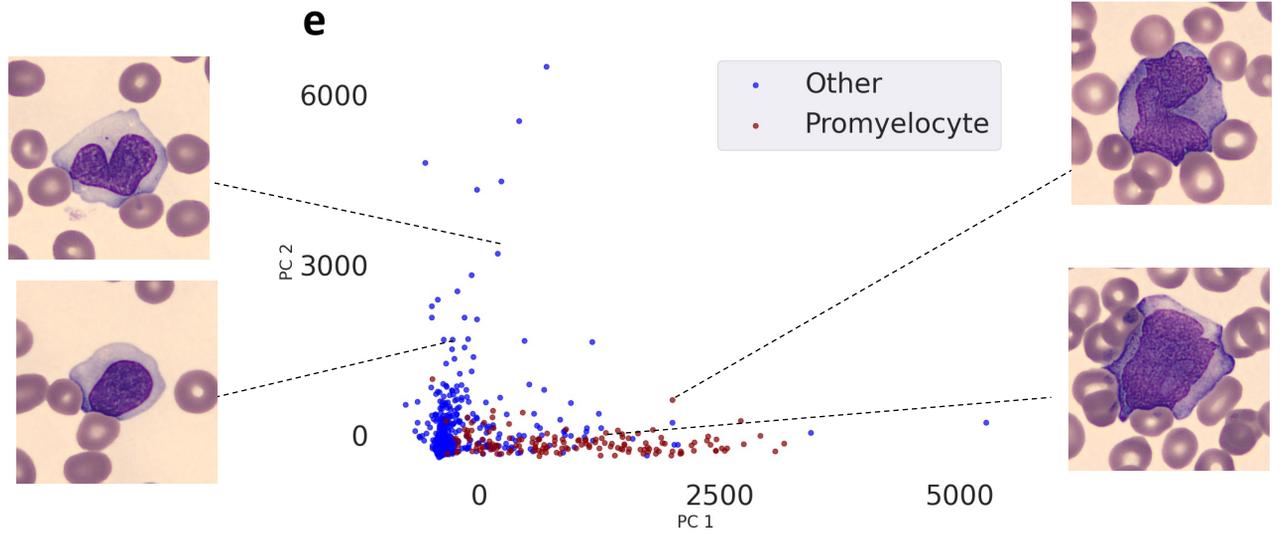

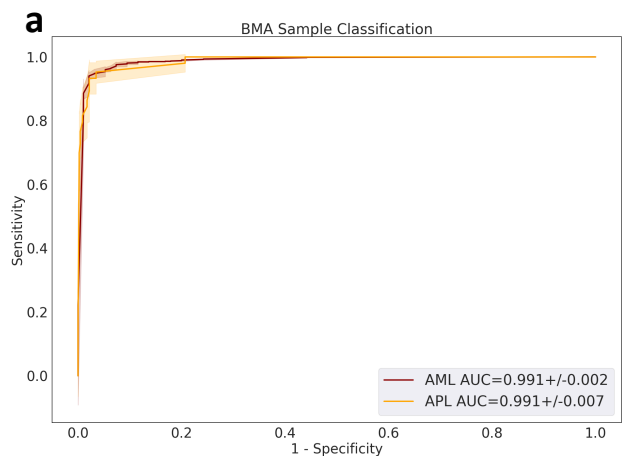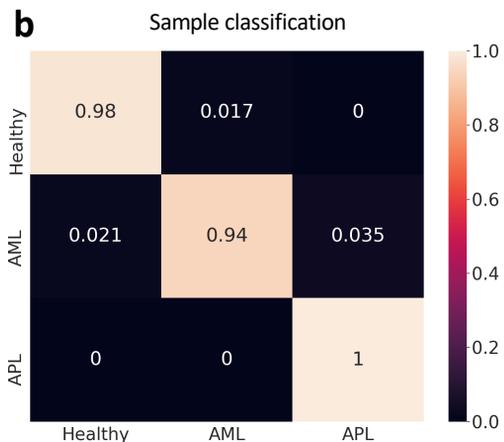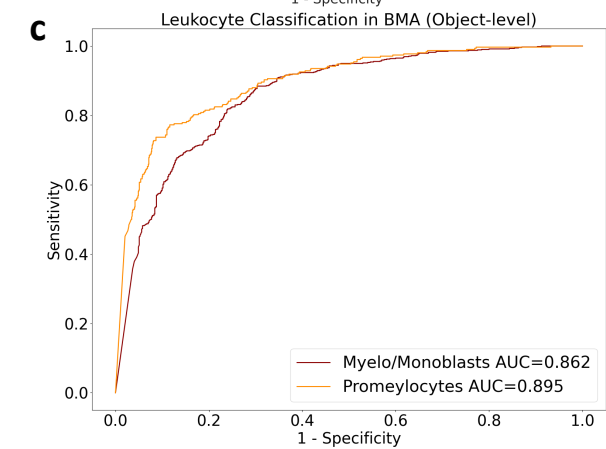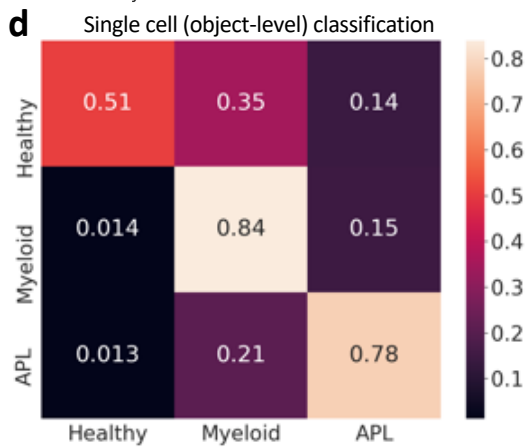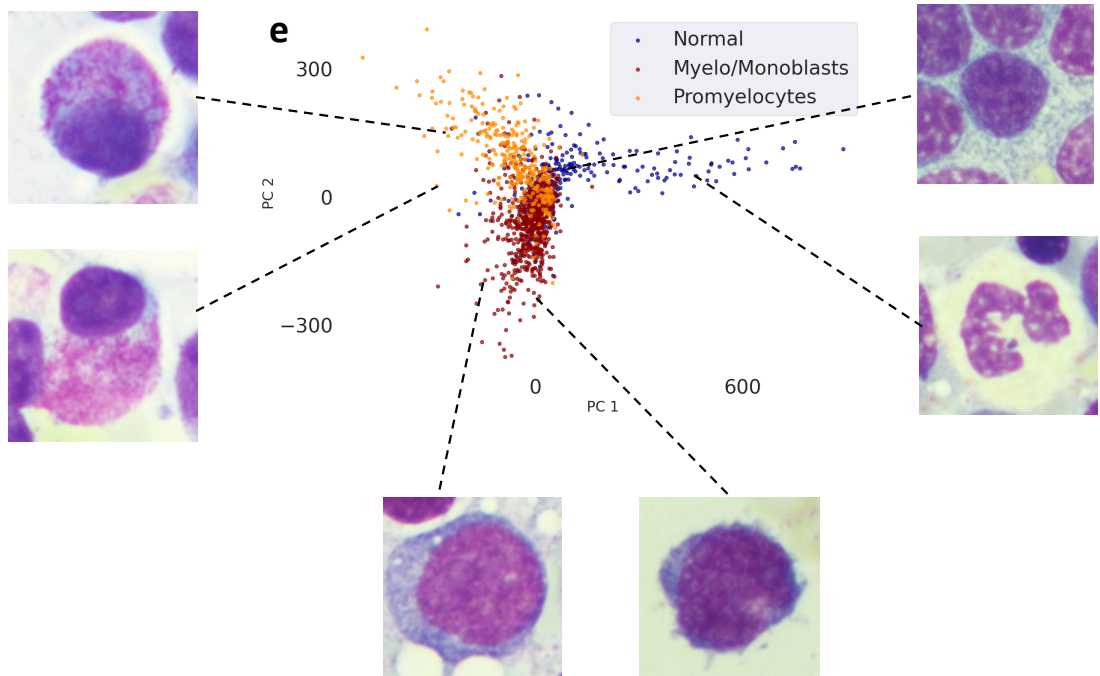